\documentclass[10pt,twocolumn,letterpaper]{article}

\usepackage{cvpr}
\usepackage{times}
\usepackage{epsfig}
\usepackage{graphicx}
\usepackage{amsmath}
\usepackage{amssymb}

\usepackage{multirow, dcolumn}
\usepackage{subcaption}
\usepackage{lipsum}
\usepackage{footnote}
\usepackage{siunitx}
\usepackage{makecell}
\usepackage{epstopdf}
\usepackage{fancyhdr}
\usepackage{xcolor}
\usepackage{gensymb}
\usepackage[normalem]{ulem}
\usepackage{soul}
\usepackage{verbatim}
\usepackage{booktabs}
\usepackage{textcomp,array,booktabs}
\usepackage[pagebackref=true,breaklinks=true,letterpaper=true,colorlinks,bookmarks=false]{hyperref}

\cvprfinalcopy 

\def\cvprPaperID{23} 

\ifcvprfinal\fi

\usepackage{cuted}
\usepackage{capt-of}
\begin{document}

\title{Residual Frames with Efficient Pseudo-3D CNN for Human Action Recognition}

\author{Jiawei Chen \quad \quad Jenson Hsiao \quad  \quad Chiu Man Ho\\
InnoPeak Technology, Palo Alto, CA, USA
{\tt\small }
%
}

\maketitle
\begin{strip}\centering
\vspace{-5ex}
\includegraphics[width=\textwidth]{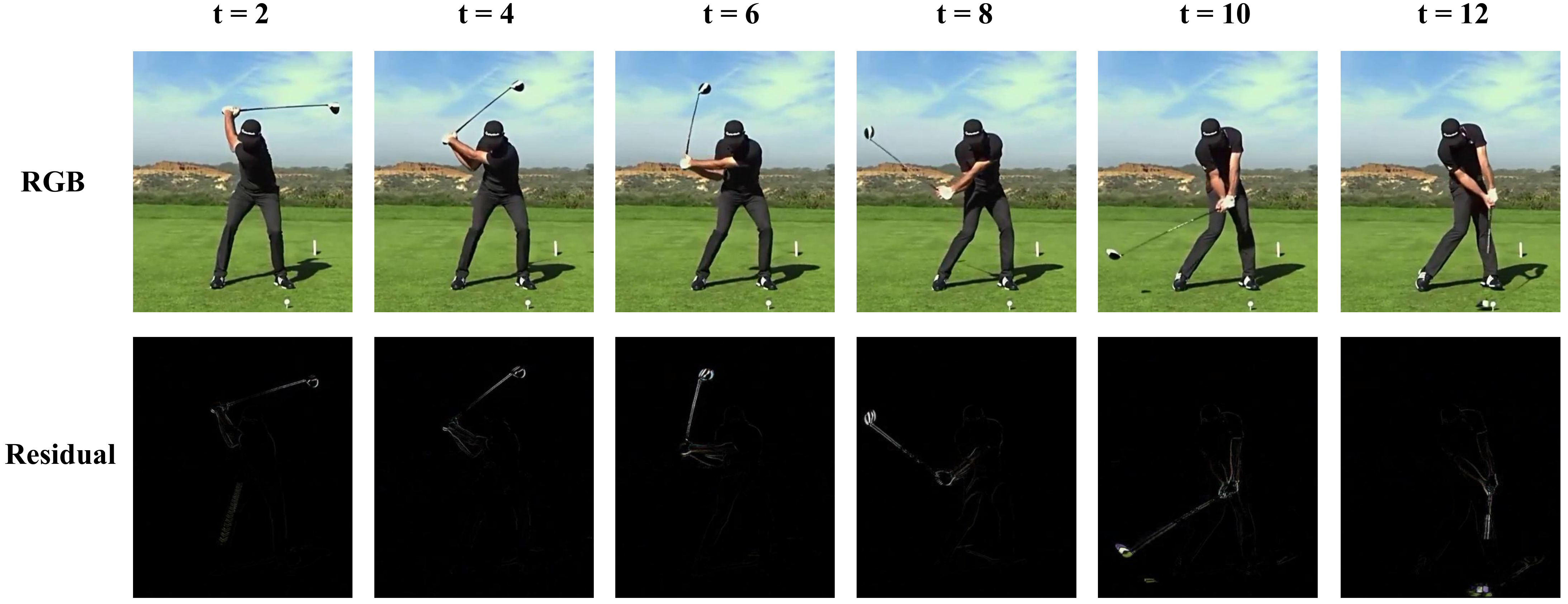}
\captionof{figure}{Examples of RGB (top) and residual (bottom) frames. The RGB frames contain rich appearance information while residual frames mainly retain salient motion cues. 
\label{fig:residual}}
\end{strip}
\begin{abstract}
Human action recognition is regarded as a key cornerstone in domains
such as surveillance or video understanding.
Despite recent progress in the development of end-to-end solutions for video-based action recognition, achieving state-of-the-art performance still requires using auxiliary hand-crafted motion representations, e.g., optical flow, which are usually computationally demanding.   
%
%
In this work, we propose to use the residual frames (i.e., differences between adjacent RGB frames) as an alternative ``lightweight''  motion representation,  which carries salient motion information and is computationally efficient. 
In addition, we develop a new pseudo-3D convolution module which decouples 3D convolution into 2D and 1D convolution.
The proposed module exploits residual information in the feature space to better structure motions, and is equipped with a
self-attention mechanism that assists to recalibrate the appearance and motion features.
Our experiment results show that using residual frames can significantly improve action recognition performance.
A thorough evaluation also verifies the efficiency and effectiveness of our proposed pseudo-3D convolution module.
\end{abstract}
\vspace{-4ex}
%

\section{Introduction}
The resurgence of convolutional neural networks (CNNs) and large-scale labeled datasets have led unprecedented advances for image classification using end-to-end trainable networks. 
However, video-based human action recognition has not yet achieved similar success based on pure CNN features.
One fundamental challenge is how to effectively model temporal information, i.e., recognizing correlation and causation through time. 
There is a classical branch of research focusing on modeling motion through hand-crafted optical flow, including histograms of flow~\cite{laptev2008learning}, motion boundary histograms~\cite{dalal2006human}, and trajectories~\cite{wang2013action}. 
In the context of deep learning, the two-stream method~\cite{chen2017semi,feichtenhofer2016convolutional,simonyan2014two,zhu2018hidden} that exploits optical flow and RGB modality in separate streams is one of the most successful frameworks.
%
%
However, it is methodologically unsatisfactory given that optical flow is computationally expensive, and two-stream methods are often not learned end-to-end jointly with the flow.
Recent studies attempt to model appearance and motion features within a single model from solely RGB modality~\cite{feichtenhofer2019slowfast,hara2018can,tran2018closer,varol2017long}.
Nevertheless, it has been shown that combining optical flow and RGB frames as input to their models can still improve performance~\cite{varol2017long}.

In this paper, we propose to use \textit{residual frames}, i.e., the differences between adjacent RGB frames, as an additional input modality to RGB data for action recognition.  
The reason is two-fold: (a) adjacent RGB frames largely share the still objects and background information, thus residual frames usually retain only motion-specific features (see Fig.~\ref{fig:residual});
(b) the computational cost of residual frames is negligible compared to other motion representations. 
In our experiments, we verify that using residual frames can yield significant improvement for action recognition. 

Following the recent trend of developing efficient 3D convolution models for video classification~\cite{lin2019tsm,qiu2017learning,tran2018closer,xie2018rethinking},
we also propose a new efficient pseudo-3D convolution module wherein the standard 3D convolution is decoupled into 2D and 1D convolution.  
To further enhance motion features, we utilize residual information in the feature space, i.e., the differences between temporally adjacent CNN features.
A novel self-attention mechanism is also proposed to recalibrate the appearance and motion features based on their significance to the end task. 

Our contributions are summarized as follows:
\begin{enumerate}
	\vspace{-1mm}
	\item 
	We propose a multi-modal approach for action recognition that utilizes both RGB and residual frames. 
    We empirically verify that using residual frames is a simple yet effective approach to improve performance. 
	\vspace{-1mm}
	\item We develop a novel and efficient pseudo-3D convolution module that involves residual features and a self-attention mechanism. We provide ablation studies to confirms the effectiveness of individual components in our proposed module. 
\end{enumerate}

\section{Methodology}
In this section, we first describe the definition of residual frames and its usage as an auxiliary modality for action recognition. Then, we introduce our efficient pseudo-3D convolution module.

\subsection{Residual Frames}
Given a video clip  $\mathbf{x} \in \mathbf{R}^{T\times H \times W \times C}$, where $T$, $H$, $W$ and $C$ denotes the clip length, height and width of each frame, and the number of channels, respectively, 
a residual frame can be formed by subtracting the reference frame $x_{t_1}$, from the desired frame $x_{t_2}$, where the step size between timestamps $t_1$ and $t_2$ is denoted as $s$.
More formally, we can define a residual frame as:
\begin{align*}
x^{res}_{(t_1, s)} =  |x_{t_1+s} - x_{t_1}|
\end{align*}
Due to nearby video frames having significant similarities in static information, a residual frame normally has little background and object appearance information but retains salient motion-specific information. Thus, residual frame is a good source for extracting motion features. 
Moreover, compared to other motion representations like optical flow, the computational cost of residual frames is notably cheaper.

In reality, actions and activities are complex and could involve different motion speed and duration.
In order to cope with such uncertainty, we can stack consecutive residual frames to form a residual clip $\mathbf{x}^{res} = \{...,x^{res}_{(i,s)},...\}$, $i =1,...,T-s$, which tends to capture fast motion in the spatial axis and slow/long-duration motion in the temporal axis. 
Thus, a residual clip is naturally suitable for 3D convolution wherein short- and long-duration motion cues can be extracted simultaneously. 

However, residual frames alone may be insufficient to solve human action recognition due to the object appearance and background scene can also provide important cues to discriminate actions, e.g., \textit{Apply Eye Makeup} and \textit{Apply Lipstick} are similar in motion but different in the location of the movement. 
%
Therefore, it is necessary to utilize both RGB and residual frames for action recognition. To this end, we develop a pseudo-3D CNN with the capability to operates both modalities. 
We will present the details in sec.~\ref{sec:pseudo-3D}
%
\subsection{Pseudo-3D Convolution Module}
\label{sec:pseudo-3D}
We propose a new pseudo-3D convolution module in which standard 3D filters are decoupled into parallel 2D spatial and 1D temporal filters.
The reason we use decoupled 3D convolution is two-fold. First, replacing 3D convolution with separable 2D and 1D convolution greatly reduces model size and computational cost, which is in line with the recent trend in the development of efficient 3D networks. 
Second, placing 2D and 1D convolution in separate pathways allows modeling appearance and motion features differently. 
In particular, we extend the idea of residual frames from \textit{pixel-level} to \textit{feature-level} when modeling motions. 
Given an output feature from 1D temporal convolution $f^{m} \in \mathbf{R}^{T' \times H' \times W' \times C'}$,  we first shift it along the temporal dimension by a stride of 1 and then generate a residual feature by subtracting the shifted feature from the original version:
\begin{align*}
f^m_{res}(t) =  |f^m(t+1) - f^m(t)|
\end{align*}
%
%
As a result, three features $\{f^s, f^m, f^m_{res}\}$ are created after the pseudo-3D convolution, where $f^s$ is the output of 2D convolution which maintains the appearance information and $f^m$, $f^m_{res}$ preserve distinctive motion structures. 

\begin{figure}[!h]
	\centering
	\includegraphics[width=0.8\linewidth]{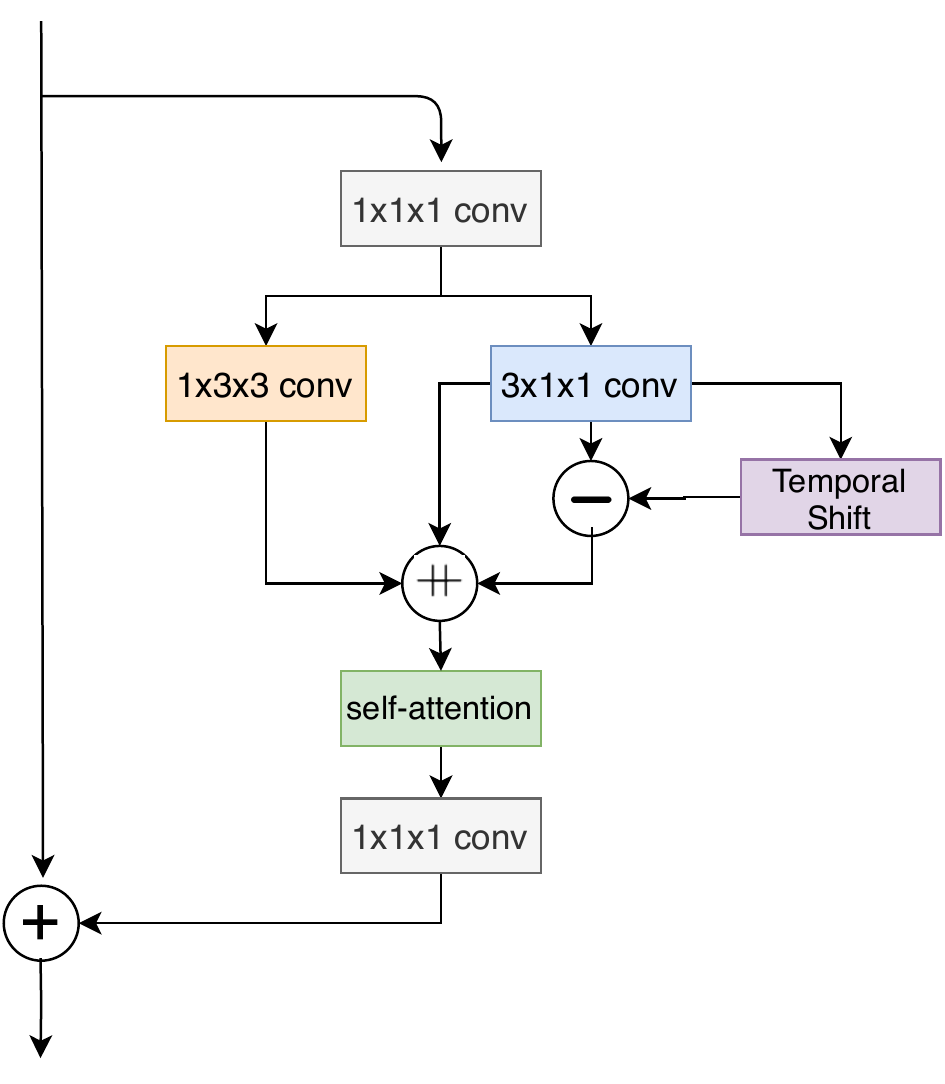}
	\caption{Proposed pseudo-3D convolution module. $+\!\!\!+$ represents concatenation.}
	\label{fig:bottleneck}
\end{figure}

To facilitate effective fusion of appearance and motion features, we propose to apply a channel self-attention mechanism to recalibrate features.
%
%
Specifically, we first concatenate output features in the channel dimension:
\begin{align*}
f = f^s+\!\!\!+ f^m +\!\!\!+ f^m_{res}, \quad f \in \mathbf{R}^{T'\times H' \times W' \times 3C'}
\end{align*}
where $+\!\!\!+$ represents concatenation.
Then, we produce an channel attention mask $M_{att}$ as:
\begin{align*}
M_{att} = \sigma(Wpool(f)+b), \quad M_{att} \in \mathbf{R}^{T'\times 1 \times 1 \times 3C'}
\end{align*}
where  $W \in \mathbf{R}^{3C' \times 3C'}$ represents a weight matrix parameterized by a one-layer neural network, $b \in \mathbf{R}^{3C'}$ is a bias term, $pool$ is a global pooling operation averaging
the dimensions of $f$ across space and time, $\sigma$ indicates the sigmoid function.
Our goal is to introduce dynamics conditioned on the input feature and reweight channels based on their significance to the end task. 
Thus, we conduct channel-wise multiplication between the input $f$ and attention mask $M_{att}$.
To further promote robustness, we adopts \textit{residual connection} in our module. 
\begin{align*}
f_{att} =  f\odot M_{att} + f
\end{align*}
%
Figure~\ref{fig:bottleneck} presents the detailed design of the proposed pseudo-3D convolution module. 
The top and bottom $1\times 1 \times 1$ convolutions are applied for reducing and restoring dimensions.

The proposed module can be integrated to any standard CNN architectures, e.g., ResNet~\cite{he2016deep}. In our experiments, we develop variants of ResNet-50 by replacing all the bottleneck block with our pseudo-3D convolution module.
In order to operate both RGB and residual frames concurrently, we modify the original data layer (the first convolutional layer) into two streams with parallel building blocks, one for each modality. The resulting features from two streams are concatenated and passed to the succeeding layer. Please refer to Table.~\ref{tbl:network} for detailed network architecture.  

\section{Experimental Evaluation}
We evaluate the performance of the proposed approach on the UCF101~\cite{soomro2012ucf101} dataset, which consists of 13,320 videos in 101 action categories. 
We preprocess each video by fixing the frame rate to 15 and resizing frames to let the short side to be 256.
%
During training, random scaling and corner cropping are utilized for data augmentation, and the
cropped region is resized to 112$\times$112 for each frame.
During testing, we uniformly sample 10 clips from each video and obtain the final prediction by averaging clip scores. 
We report video-level top-1 and top-5 accuracy on the validation set of split-1 for all experiments. 
It is worthwhile to mention that state-of-the-art performance on UCF101 is achieved by using deep models that have been pretrained on large-scale video datasets~\cite{carreira2017quo,lin2019tsm}.
However, we train all the models from scratch since pushing the performance limit is out of our current scope.  
\begin{table}
\caption{The proposed network architecture. The dimensions of filters are denoted by $\{ T\times S^2, C\}$ for temporal, spatial and channel sizes. Please note we omit the attention layer for simplicity.}
\label{tbl:network}
\centering
\resizebox{\columnwidth}{!}{
\begin{tabular}{c|c|c}
    \hline
    Stage & Filters &  Output size $T\times S^2$ \\
    \hline 
    raw clip &
    \makecell{-}
    &
    \makecell{\textit{RGB}: $32\times112^2$ \\ \textit{Residual}: $32\times112^2$}\\
    \hline
    \multirow{2}{*}{conv1} & \textit{RGB}: [$1 \times 7^2$, 64], [$3 \times 1^2$, 64] & \textit{RGB:} $32\times 56^2$\\
    & \textit{Residual}: [$1 \times 7^2$, 64], [$3 \times 1^2$, 64] & \textit{Residual:} $32\times 56^2$ \\
    \hline
    \makecell{res2} 
    &
    $\Bigg[$
    \makecell{
        $1\times1^2$, 64\\
        \small[$1\times3^2$,64\small], $\small[ 3\times 1^2$, 64 \small]\\
        $1\times1^2$, 64
    }
    $\Bigg]\times3$

    & $32\times56^2$\\
    \hline
    \makecell{res3} 
    &
    $\Bigg[$
    \makecell{
        $1\times1^2$, 128\\
        \small[$1\times3^2$,128\small], $\small[ 3\times 1^2$, 128 \small]\\
        $1\times1^2$, 128
    }
    $\Bigg]\times4$
    & $32\times28^2$\\
    \hline
    
    \makecell{res4} 
    &
    $\Bigg[$
    \makecell{
        $1\times1^2$, 256\\
        \small[$1\times3^2$,256\small], $\small[ 3\times 1^2$, 256\small]\\
        $1\times1^2$, 256
    }
    $\Bigg]\times6$
    & $32\times14^2$\\
    \hline
    
    \makecell{res5} 
    &
    $\Bigg[$
    \makecell{
        $1\times1^2$, 512\\
        \small[$1\times3^2$,512\small], $\small[ 3\times 1^2$, 512 \small]\\
        $1\times1^2$, 512
    }
    $\Bigg]\times3$
    & $32\times7^2$\\
    \hline
    pool & $32\times 7^2$ & $1\times 1 \times 1$\\
    \hline
    fc1 & $1\times 1^2$, 2048& $1\times 1 \times 1$\\
    \hline
    fc2 & $1\times 1^2$, $\#$classes & $1\times 1 \times 1$\\
    \hline
\end{tabular}
}
\end{table}

In specific, we first conduct an evaluation of the effectiveness of different data modalities by training action classifiers using solely RGB frames, residual frames and a combined input, respectively. 
We also study the impact of the step size of residual frames for action recognition. 
%
Finally, we perform ablation studies to investigate the effectiveness of individual components in the proposed pseudo-3D convolution module. 

%

\noindent
\textbf{Performance comparisons for different data modalities.}
%
Table.~\ref{tbl:input_results} shows the action recognition performance of various combinations of input modality and network architecture on UCF101. 
Note for experiments using solely RGB or residual frames, we keep only one stream in the data layer (\textit{conv1}) but double the number of channels for a fair comparison. 
We first observe that using solely residual frames outperforms solely RGB frames by $\sim 3\%$ in top-1 and top-5 accuracy. 
It indicates residual frames indeed contain salient motion information which is important for action recognition. 
When leveraging both RGB and residual frames,  the top-1 accuracy is further increased by $2.6\%$ ($83.0\%$ to $85.6\%$), which suggests the two data modalities maintain complementary information. 
Remarkably, we also observe that using our pseudo-3D convolution module significantly reduces computing flops (from 163G to 30G)
compared to the case of using standard 3D convolution, while still providing better performance.
%
%
\begin{table}[!thb]
\vglue -0cm
\caption{Performance comparisons for different input modalities.}
\label{tbl:input_results}
\centering
\setlength\extrarowheight{1pt}
\resizebox{1\columnwidth}{!}{%
\begin{tabular}{ c c c c c }
\Xhline{1pt}
\textbf{Method} & \multirow{1}{*}{\textbf{Input modality }}	
&   \multicolumn{1}{c}{\textbf{Val top-1 }} 
&   \multicolumn{1}{c}{\textbf{Val top-5}} & \textbf{GFLOPs} \\
\hline
P.3D ResNet50 & RGB & $80.4\%$ &  $95.0\%$ & 28\\
\hline
P.3D ResNet50 & Residual ($s=1$) & $83.0\%$ & $98.3\%$ & 28\\
\hline
\multirow{2}{*}{3D ResNet50}  & RGB +  & \multirow{2}{*}{$85.0\%$} &  \multirow{2}{*}{$99.1\%$} & \multirow{2}{*}{163}\\
& Residual ($s=1$)& & &\\
\hline
\multirow{2}{*}{P.3D ResNet50}  & RGB +  & \multirow{2}{*}{$85.6\%$} &  \multirow{2}{*}{$99.2\%$} & \multirow{2}{*}{30}\\
& Residual ($s=1$)& & &\\
\Xhline{1pt}
\end{tabular}}
\vglue -0cm
\end{table}

\noindent
\textbf{Impact of step size $s$.}
When generating residual frames, we can change the step size $s$ to capture motion features at different time scales. 
However, it is unclear what the optimal step size is for the action recognition task. 
Therefore, we conducted study on the impact of the step size and show the results in Table.~\ref{tbl:stepsize_results}.
We experimented three settings where the input data is solely residual frames with step size $s =1,2, 4$, respectively. 
Interestingly, the classification accuracy decreases with the increase of step size. 
We suspect it is because motion will cause spatial displacements for the same objects between two frames, and it may cause a mismatch in motion representations when using a large step size.

\begin{table}[!thb]
\vglue -0cm
\caption{Performance comparisons for different residual frame step size $s$.}
\vglue -0.30cm
\label{tbl:stepsize_results}
\centering
\setlength\extrarowheight{1pt}
\resizebox{0.8\columnwidth}{!}{%
\begin{tabular}{ c c c }
\Xhline{1pt}
\multirow{1}{*}{\textbf{Step size}}	
&   \multicolumn{1}{c}{\textbf{Top-1 accuracy}} 
&   \multicolumn{1}{c}{\textbf{Top-5 accuracy}} \\
\hline		
$s=1$ & $83.0\%$ &  $98.3\%$\\
\hline
$s=2$ & $82.7\%$ & $97.1\%$ \\
\hline
$s=4$ & $80.2\%$ &  $96.0\%$\\
\Xhline{1pt}
\end{tabular}}
\vglue -0cm
\end{table}

\noindent
\textbf{Ablation studies.}
We perform ablation studies to verify the effectiveness of different components in our proposed pseudo-3D convolution module. Without loss of generality, the models are trained with a combined input of RGB and residual frames ($s=1$). 
As shown in Table.~\ref{tbl:ablation_results}, removing the self-attention mechanism leads to a $1.8\%$ drop in top-1 accuracy ($85.6\%$ to $83.8\%$). 
Meanwhile, the performance is also reduced from $85.6\%$ to $83.5\%$ when ignoring the residual information in the feature space. 
If the self-attention mechanism and residual features are eliminated at the same time, the top-1 accuracy will be further reduced to $82.1\%$. 
These results confirm the self-attention mechanism and residual features are effective to improve the performance of action recognition.


\begin{table}[!thb]
\vglue -0cm
\caption{Performance comparisons for various pseudo-3D module settings.}
\vglue -0.30cm
\label{tbl:ablation_results}
\centering
\setlength\extrarowheight{1pt}
\resizebox{1\columnwidth}{!}{%
\begin{tabular}{ c c c }
\Xhline{1pt}
\multirow{1}{*}{\textbf{Method}}	
&   \multicolumn{1}{c}{\textbf{Top-1 accuracy}} 
&   \multicolumn{1}{c}{\textbf{Top-5 accuracy}} \\
\hline
P.3D module w.o. & \multirow{2}{*}{$82.1\%$}& \multirow{2}{*}{$97.9\%$}\\
self attention $\&$ residual feature & & \\
\hline
P.3D module w.o. self-attention & $83.8\%$ &  $98.4\%$\\
\hline
P.3D module w.o. residual feature & $83.5\%$ & $98.1\%$ \\
\hline
Pseudo 3D module & $85.6\%$ &  $99.2\%$\\
\Xhline{1pt}
\end{tabular}}
\vglue -0cm
\end{table}

\section{Conclusion}
In this paper, we propose a multi-modal framework that exploits both RGB and residual frames for human action recognition. We empirically confirm the benefit of using residual frames ($5.6\%$ increase on top-1 accuracy) and our study shows that using small step size for residual frames  will lead to better performance. 
One additional contribution of this paper is the development of a novel and efficient pseudo-3D convolution module that involves residual features and a self-attention mechanism.
Quantitative results show that the proposed module can significantly reduce computational cost without compromising performance. 

{\small
\bibliographystyle{ieee_fullname}
\bibliography{egbib}
}

\end{document}